\documentclass{article}
\usepackage{spconf}
\usepackage[table]{xcolor}
\usepackage{times}
\usepackage{latexsym}
\usepackage{graphicx}
\usepackage{amsmath}
\usepackage{url}
\usepackage{multirow}
\usepackage{dialogue}
\usepackage{caption}
\usepackage{dingbat}
\pdfoutput=1

\title{User Modeling for Task Oriented Dialogues}

\name{Izzeddin G\"{u}r$^{1,2}$, Dilek Hakkani-T\"{u}r$^3$\sthanks{Work performed while at Google AI}, Gokhan  T\"{u}r$^{4\small{*}}$, Pararth Shah$^{5\small{*}}$}
\address{
	$^1$University of California, Santa Barbara, CA, USA\\%
	$^2$Google AI, Mountain View, CA, USA\\%
    $^3$Amazon Alexa AI, Mountain View, CA, USA\\%
	$^4$Uber AI Labs, San Francisco, CA, USA\\%
    $^5$Facebook Conversational AI, Menlo Park, CA, USA\\%
	izzeddingur@cs.ucsb.edu, \{dilek, gokhan.tur\}@ieee.org, pararths@fb.com
}

\begin{document}
\ninept
\maketitle
\begin{abstract}
	We introduce end-to-end neural network based models for simulating users of task-oriented dialogue systems. User simulation in dialogue systems is crucial from two different perspectives: (i) automatic evaluation of different dialogue models, and (ii) training task-oriented dialogue systems. 
	We design a hierarchical sequence-to-sequence model that first encodes the initial user goal and system turns into fixed length representations using Recurrent Neural Networks (RNN). It then encodes the dialogue history using another RNN layer. At each turn, user responses are decoded from the hidden representations of the dialogue level RNN. This hierarchical user simulator (HUS) approach allows the model to capture undiscovered parts of the user goal without the need of an explicit dialogue state tracking. We further develop several variants by utilizing a latent variable model to inject random variations into user responses to promote diversity in simulated user responses and a novel goal regularization mechanism to penalize divergence of user responses from the initial user goal. We evaluate the proposed models on movie ticket booking domain by systematically interacting each user simulator with various dialogue system policies trained with different objectives and users.
\end{abstract}

\section{Introduction}

Task oriented dialogue systems aim to help users accomplish their goals via interacting with them in natural language. Previous work on task oriented dialogue systems can be categorized into two main paradigms: (i) pipelined architectures with modularly connected components that are trained independently \cite[among others]{young00,voicetone,danbohus:2009}, and (ii) end-to-end architectures where components can be jointly trained~\cite[among others]{wen16,tonyzhao:2016,xuesongyang:2017,bingliu:2017}. End-to-end training aims to prevent transfer of errors through the component pipeline and enable continuous training as new data becomes available, even when annotations are not available for each component. Investigation of end-to-end approaches for user simulation is motivated by the possibility of continuous training of user models from unannotated human conversations.
End-to-end training has also been shown to result in higher task success rates and shorter dialogues~\cite [among others]{bingliu:2017}.
Typically, these models are trained with labeled dialogues from various sources, including simulated dialogues enriched with crowd realizations. 

\begin{figure}[t]
	\textbf{User Goal: } \textit{date=}Friday, \textit{num\_tickets=}2, \textit{theatre\_name=}DontCare, \textit{movie=}Sully, \textit{time=}DontCare \\ \newline
	\scalebox{0.8}{
		\begin{tabular}{lll}
			0 - & \textsf{SYSTEM} & Hi, how can I help you?\\
			&&{\tt {\small greeting()}} \\
			& \textsf{USER} & Hi, I'd like to buy tickets to see the Sully movie.\\
			&&{\tt {\small greeting() intent(buy\_movie\_tickets)}} \\ &&{\tt {\small inform(movie=Sully) }}\\
			1 - & \textsf{SYSTEM} & You wanna see Sully. How many tickets would you need?\\
			&&{\tt {\small confirm(movie=Sully) request(num\_tickets) }}\\
			& \textsf{USER} & 2 tickets please. \\
			&& {\tt {\small inform(num\_tickets=2)}} \\
			&&...
		\end{tabular}
		\vspace*{-1ex}
	}
	\caption{An example dialogue from the movie ticket booking domain. Each dialogue has an initial user goal randomly assigned at the beginning. 
	} 
	\label{fig:ex_dialogue}
	\vspace*{-2.5ex}
\end{figure}

Recently, sequence-to-sequence (seq2seq) user simulators for dialogue systems have been built for training dialogue policy models using reinforcement learning~\cite[among others]{asri16,Baolin,crook2017sequence}.
Even though these simulators can generate successful conversations, they require annotated dialogue states and lack necessary mechanisms to produce diverse responses which limits their capability to evaluate different dialogue systems.

Besides training end-to-end dialogue systems, modeling user behavior has other use cases. Google Duplex demo\footnote{https://www.youtube.com/watch?v=bd1mEm2Fy08} is a good example of machines mimicking humans for achieving certain tasks. Furthermore, these simulators can be used to evaluate the alternative approaches for a task oriented dialogue system. Most systems are evaluated on fixed testing corpora, a very limited way in its ability to test if a system can accurately interact with various different users. Another way of evaluating task oriented dialogue systems is with crowd sourcing, where one can more accurately capture successful user interactions. But, its costly and time consuming nature only allows a limited set of dialogues to be generated. Furthermore, task oriented dialogue systems have various model decisions that are infeasible to make using crowd sourcing. A scalable and accurate evaluation criteria that can compare multiple systems in a similar set-up is critical to ameliorate the aforementioned bottlenecks. 

In this paper, we propose several end-to-end user simulation models that aim to mimic realistic and diverse user behaviors.  Our user simulators aim to enable side-by-side comparisons of dialogue systems or their components across a given set of user goals and dialogue scenarios.  Another goal of our approach is to enable training of dialogue policies via supervised and reinforcement learning using a diverse set of interactions.

Inspired by recent conversational models \cite{serban16}, we develop a hierarchical seq2seq user simulator (HUS) that implicitly tracks user goal over multiple turns. HUS first encodes a user goal and each system turn observed in the dialogue history into vector representations. Initialized from the user goal vector, a higher level encoder generates a dialogue history representation using system turn vectors as input at each time step. Finally, the simulated user turns are decoded from this dialogue history representation.

While HUS can generate successful dialogues, it will always generate the same simulated user turns given the same user goal and system turns. To induce human-like variations and generate a richer set of user turns, we propose a variational framework where an unobserved latent variable generates user turns.
Furthermore, with HUS, user turns and user goals are related only through a user turn decoder, resulting in repetition of previously specified information throughout the interaction.
We introduce a new goal regularization approach where a penalization term that aims to maximize the overlap of tokens between the user turns and the user goals is added to the final loss.
The proposed user simulators work at the dialogue act level and generate responses in the form of acts and associated slot and value pairs (a sample dialogue is shown in Figure \ref{fig:ex_dialogue}).
Final outputs are then converted into natural language utterances using a template-based natural language generation approach.

We evaluate our models on a \textit{movie ticket booking} domain by systematically interacting each user simulator with multiple system policies trained with different objectives and architectures. We show that a reinforcement learning based policy is more robust to random variations in user behavior, while a supervised model suffers from a considerable drop in accuracy.
Our goal regularization approach generates significantly shorter dialogues (i.e. avoiding loops around user or system misunderstandings) across all system policies and shows higher task completion rates. Finally, human evaluations show high naturalness scores for all proposed approaches.

\section{Related Work}
The proposed model architectures in our work are inspired by hierarchical deep learning models studied in \cite{sordoni15,serban16,jiwei15} which are also shown to perform better than plain seq2seq and language models on dialogue generation tasks.
Variational approaches such as Variational Autoencoders are used for unsupervised or semi-supervised model training \cite{kingma13,fabius14} that improves the final performance metrics as well as the diversity of the generated outputs.

Agenda based user simulations have been investigated in task oriented dialogues for model training \cite{schatzmann07}. 
Supervised learning approaches using linear models \cite{georgial2015learningus,georgial2016usersf} as well as hidden markov models \cite{cuayahuitl2005hmm} have also been proposed for simulating users.
Recently, seq2seq neural network models are proposed for user simulation \cite[among others]{asri16,Baolin,kreyssig2018sim} that utilize additional state tracking signals and dialogue turns are encoded more coarsely. Kreyssig et al. \cite{kreyssig2018sim} further showed that plain seq2seq models with state tracking signals at each turn can outperform agenda-based user simulators on several metrics including success rate. 
Another work that also uses seq2seq models for user simulation is~\cite{crook2017sequence}, where seq2seq models are compared with language modeling based approaches for generating user turns in natural language. They encode the context (previous user turn) and current system turn using two independent encoders and concatenate the output vectors. Simulated user turn is then decoded from this final vector. However, we propose an end-to-end hierarchical seq2seq approach that can encode the entire dialogue history without any feature extraction and does not require any external state tracking annotations. Furthermore, we introduce several novel variants using variational approaches and goal regularization that improves the dialogue metrics and allows a more thorough comparison with different system policies.

\section{Problem Description}
\vspace*{-2
	ex}
Following the recent progress on end-to-end supervised dialogue models, we consider the inverse problem of generating a user turn $U_t$ given a user goal $C$ and the history of system turns $S_1,S_2,\cdots,S_t$.
A user goal is a set of slot-value pairs (ex. \textit{time: 12pm}) with a predefined user personality (e.g., \textit{aggressive, cooperative}) that defines the sampling distribution when generating user turns.
Similarly, system and user turns are sets of actions where each action has a dialogue act (ex. \textit{inform}) and a set of corresponding slot-value pairs, e.g., \textit{inform(time=12pm, theatre="AMC Theatre")}.

We first generate a more coarse level representation of each input by replacing the value of each slot with one of the following values: \{{\tt Requested, DontCare, ValueInGoal, ValueContradictsGoal, Other} \}. If the value of a slot is requested by system, we replace those slot values with {\tt Requested}.
If the value of a slot appears in or contradicts the user goal, we replace those values with {\tt ValueInGoal} or {\tt ValueContradicts\-Goal}, respectively. If the value of a slot in the user goal is flexible, meaning the user is open to accept any offered value, we replace the value of the corresponding slot with {\tt DontCare} in each input. Finally, we replace other values with {\tt Other}. During testing, based on the coarse value, we sample an actual value either from the user goal, system turn, or from the knowledge base. Inspired by the success of recent approaches to structured data, we next linearize each input (user scenario, system and user turns) following \cite{vinyals14} to generate token sequences. As an example, we replace the actual movie name \textit{Sully} in system Turn-1 from Figure \ref{fig:ex_dialogue} with {\tt ValueInGoal} and convert the sequence of actions into a sequence of tokens as \newline {\tt \small "confirm", "(", "movie=ValueInGoal", ")",\\ "request", "(" "num\_tickets", ")"}.

We now describe our problem more formally. Given a user goal sequence $C$ and the history of system turn sequences $S_t$ at each turn, our task is to generate a new sequence of tokens that match the correct user turn sequence $U_t=\{u_{t1},u_{t2},\cdots,u_{tN_U}\}$ at turn $t$. Here, a successful user turn might require blending information from the user goal, and history of system and user turns. Furthermore, we do not assume that a supervised signal for dialogue state tracking is given.

\begin{figure*}[t]	
	\centering
	\includegraphics[width=0.7\linewidth]{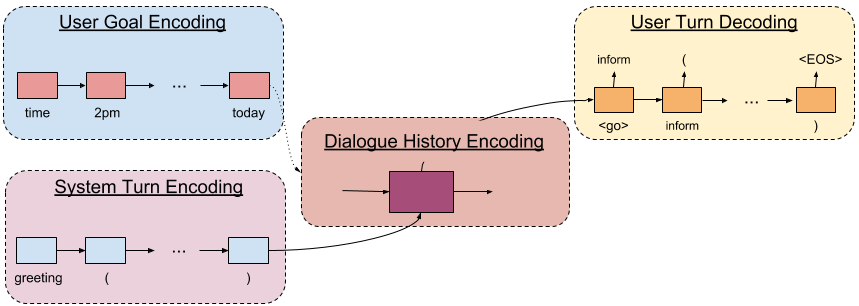}  
	\vspace*{-1ex}
	\caption{HUS model: Boxes are RNN cells, colors indicate parameter sharing.}
	\label{fig:hus} 
	\vspace*{-2ex}
\end{figure*}

\section{Hierarchical Seq2Seq Models for User Simulation}
We now present our supervised models, in Figure~\ref{fig:hus}, and describe their operations in a fully end-to-end setting. Our baseline model is a hierarchical seq2seq neural network that operates on individual system and user turns as well as at the overall dialogue level. Given a $(C, S_{t}, U_{t})$ triplet, the model first encodes user goal $C$ and the system turn dialogue act at turn $t$, $S_{t}$, into vector space representations in parallel with decoupled Recurrent Neural Networks (RNN). Another RNN is initialized from the goal representation and encodes the sequence of system turn representations into a hidden vector at turn $t$. Finally, simulated user response in dialogue act and arguments at turn $t$, $U_t$, is decoded from this hidden vector using an RNN sequence decoder. We next discuss the shortcomings of this model and propose several model improvements that are unsupervised in the sense that no additional supervised data is required.

\subsection{Preliminaries and Notation}
Each token $w$ comes from a vocabulary $V$ and is associated with a vector $e_w$. $E^C, E^{S_t}$ and $E^{U_t}$ denotes the sequence of token embedding vectors for user goal, system and user turns at turn $t$, respectively.
Our models utilize the power of RNN encoders and decoders where we utilize GRU units \cite{cho14}.
We refer to the average of hidden states of a turn level RNN encoder as the encoding of a sequence.

\subsection{Hierarchical Seq2Seq User Simulation (HUS)}
The core of our model is a hierarchical seq2seq neural network that first encodes user goal and system turns and generates user turns from dialogue level representations.
\newline \newline
\textbf{Encoding User Scenario and System Turns}
Using a RNN ($Enc$), we encode user goal as
\begin{equation}
h^C = Enc(e^C;\theta_C)
\end{equation}
\noindent where $h^C$ is the encoding and $\theta_C$ represents the parameters of goal encoder. We employ another RNN to encode each system turn as
\begin{equation}
h^S_i = Enc(e^{S_i};\theta_S)
\end{equation}
\noindent where $h^S_i$ is encoding and $\theta_S$ represents the parameters of system turn encoder. $\theta_S$ is shared for every system turn encoder and is decoupled from $\theta_C$.

\textbf{Encoding Dialogue History}
Given the sequence of system turn encodings, $h^S_0,h^S_1,\cdots,h^S_t$, we employ another RNN conditioned on the user goal encoding to encode dialogue history as
\begin{align}
	h^D_0 &= h^C \\
	h^D_t &= Enc(\{h^S_i\}_{i=1,\cdots, t};\theta_D)
\end{align} 
\noindent where $h^D_0$ is the initial hidden state, $h^D_t$ is the history encoding, and $\theta_D$ represents the parameters of the dialogue level RNN. Ideally, the dialogue level RNN should keep track of the user goal and generate meaningful user turns.

\textbf{Decoding User Turns}
Given the history encoding at turn $t$, an RNN decoder ($Dec$) generates the user turn token sequence:
\begin{align}
	h^U_0 &= h^D_t \\
	h^U_{t,i} &= Dec(h^U_{t-1,i};\omega_U) \\
	P(U^*_{t,i} &= w_j) \approx exp(e_j^T (W_U h^U_{t,i} + b_U)) \\
	U^*_{t,i} &= argmax_j (P(U^*_{t,i} = w_j))
\end{align}
\noindent where $h^U_0$ is the initial hidden state and $H^U_{t,i}$ is the hidden vector at turn $t$ and timestep $i$. $U^*_t$ is the sequence of user turn tokens generated. $W_U,b_U$ and $\omega_U$ are the parameters of the user turn decoder. The training objective is to minimize the cross-entropy error $L_{crossent}$ between the candidate sequence $U^*$ and the correct user turn $U$.

\textbf{Incorporating Dialogue Length}
During training, we observe that sampling probabilities of different user personalities randomly generate shorter or longer dialogues which is not captured by HUS.
To overcome this problem, we append the length of the corresponding dialogue to system turn encodings at each turn during training.
During testing, we randomly sample a dialogue length from the following normal distribution $\mathcal{N}(5, 2)$.
For the subsequent sections, we always incorporate dialogue length in our models.

An alternative model would also incorporate the previous user turn more explicitly, such as by encoding user turns and conditioning dialogue history encoder on user turn encodings; however we observe no significant benefit.
One plausible reason is that, user turns are decoded from history encodings which are implicitly conditioned on previous user turns via previous history encodings.

\begin{figure*}[t]	
	\centering
	\includegraphics[width=0.7\linewidth]{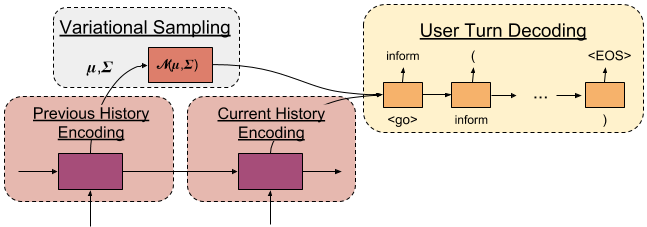} 
	\vspace*{-1ex}
	\caption{VHUS model: A variational sampling step before user turn decoder is proposed. Boxes are RNN cells, colors indicate parameter sharing. }
	\label{fig:vhus} 
	\vspace*{-3ex}
\end{figure*}

\subsection{Variational HUS}
The HUS model will generate exactly the same user turns given the same user goal and dialogue history. However, to explore the robustness of a system policy and model different types of users, we need a model that can generate more diverse and at the same time meaningful user turns. Here, we propose a novel variational hierarchical seq2seq user simulator (VHUS) where an unobserved latent random variable generates the user turn sequence (Figure \ref{fig:vhus}). The encoders are exactly the same as HUS, but the hidden state $h^D_t$ is not directly passed to the decoder. Instead, it is first concatenated with a latent vector generated from a Gaussian distribution with a diagonal covariance matrix and then passed to the decoder.

More formally, given the sequence of dialogue representations $h^D_0,h^D_1,\cdots,h^D_t$, we learn a prior Gaussian distribution $\mathcal{N}(z | \mu_x, \Sigma_x)$ using the previous dialogue history: The mean and covariance is estimated as follows
\begin{align}
	\mu_x &= W_{\mu} h^D_{t-1} + b_{\mu} \\
	\Sigma_x &= W_{\Sigma} h^D_{t-1} + b_{\Sigma}
\end{align}
\noindent where $W_{\mu}$, $W_{\Sigma}$, $b_{\mu}$, and $b_{\Sigma}$ are new parameters for prior distribution. The decoder is then initialized with a single vector $\hat{h}^D_t = FC([h^D_t;z_x])$ where $[.]$ is vector concatenation, $FC$ is a single layer neural network, and $z_x$ is sampled from the prior distribution, $z_x \sim \mathcal{N}(z | \mu_x, \Sigma_x)$. We also learn a posterior Gaussian distribution using the current dialogue history as
\begin{align}
	\mu_y &= \tilde{W}_{\mu} h^D_{t} + \tilde{b}_{\mu} \\
	\Sigma_y &= \tilde{W}_{\Sigma} h^D_{t} + \tilde{b}_{\Sigma}
\end{align}
\noindent where $\tilde{W}_{\mu}$, $\tilde{W}_{\Sigma}$, $\tilde{b}_{\mu}$, and $\tilde{b}_{\Sigma}$ are new parameters for posterior distribution. We add the Kullback-Leibler divergence between trained prior and posterior distribution, i.e., 
\[L_{var} = \alpha KL(\mathcal{N}(z | \mu_x, \Sigma_x) | \mathcal{N}(z | \mu_y, \Sigma_y)),\]
\noindent
to the cross-entropy loss, where $\alpha$ is a balancing parameter between the two losses.

\begin{figure*}[t]	
	\centering
	\includegraphics[width=0.75\linewidth]{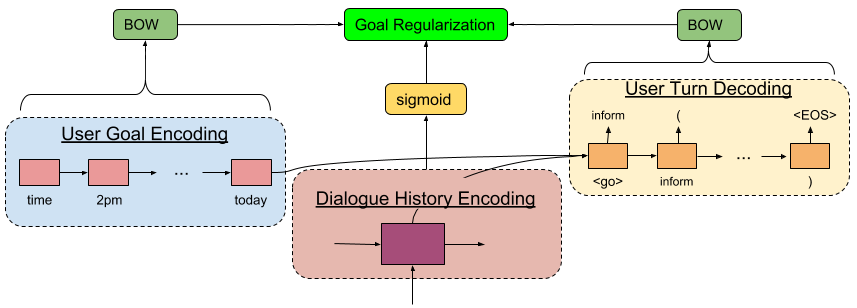}  
	\vspace*{-1ex}
	\caption{VHUSReg model: Divergence between user turns and user goal is regularized. Boxes are RNN cells, colors indicate parameter sharing. }
	\label{fig:vhusreg} 
	\vspace*{-2ex}
\end{figure*}

Our main intuition is that, when the decoder is conditioned on a noisy history, it will generate a slightly different user turn. By penalizing the KL divergence between prior and posterior distributions, we control the level of noise by ensuring that the previous and current histories are consistent.

\subsection{Goal Regularization}
HUS and VHUS models learn the relationship between a user turn and a user goal only through the decoder loss which might generate considerably longer dialogues when user turns diverge from the initial user goal. Here, we introduce a new goal regularization approach called VHUSReg to eliminate the aforementioned problem (Figure \ref{fig:vhusreg}).
Our turn level encoders are exactly the same as in HUS, but instead of initializing the dialogue level RNN with user goal representation, we initialize it with zero vector and condition the decoder on user goal more directly. Furthermore, we penalize the divergence between dialogue level representations and user goal to enforce a more direct correspondence.

More formally, given the sequence of system turn encodings, we generate dialogue representation as
\begin{equation}
h^D_t = Enc(\{h^S_i\}_{i=1,\cdots, t};\theta_D)
\end{equation}
\noindent where the encoder is initialized with zeros.
Next, we generate a new dialogue representation as
\begin{equation}
\hat{h}^D_t = FC([h^D_t; h^C])
\end{equation}
\noindent where new output is generated by blending user goal and old dialogue representation. Similarly, an RNN decoder is used to generate the user turn token sequence from new dialogue representation.

To regularize the divergence of user turn and user goal, we minimize the discrepancy between user turn and user goal tokens conditioned on the current system turn.
We first generate a bag-of-words approximation of current user turn and current system turn:
\begin{align}
	b^D_t &= FC(h^D_t) \\
	b^S_t &= FC(h^S_t)
\end{align}
where $FC$ is a single layer neural network with sigmoid activation function.
Following, we build a new bag-of-words representation for current user turn conditioned on the current system turn:
\begin{equation}
b_t^u = FC([b^D_t;b^S_t])
\end{equation}
Next, we introduce a new loss to minimize the divergence between initial user goal and user turn tokens by ensuring that each bag-of-words approximation is accurate :
\begin{multline}
	L_{reg} = || b_{t}^u - BOW(C)|| + || b^D_t - BOW(U_t)|| + \\ || b^S_t - BOW(S_t)||
\end{multline}
\noindent where $BOW(x)$ is a function that outputs a bag-of-words vector for a set of tokens $x$, for example we use $BOW(C)$ to generate a bag-of-words vector for initial user goal.
By minimizing the discrepancy between the bag-of-words representations in each term, we align the tokens in user goal and user turn.

\subsection{Variational HUS with Goal Regularization}
Finally, we combine both approaches in a hybrid framework and sum the three losses:
\begin{equation}
L = L_{crossent} + L_{var} + L_{reg}
\end{equation}

\section{Experimental Results and Discussion}
\vspace*{-2ex}
We present our experimental results by systematically interacting each user simulator model with two different system policies, trained with supervised learning and reinforcement learning. We use a dataset from \textit{movie ticket booking} domain and use several metrics to evaluate and compare our models, including task completion rates~\cite{pietquin2013survey} and dialogue length. We use template-based NLG to produce utterances for user and system action sequences, where we sample a template from a set of templates written by crowd workers for each dialogue act and the set of arguments that are possible for this application.

\subsection{System Policy Models}

For comparison, we train two state-of-the-art end-to-end system policy models: (i) supervised policy, and (ii) reinforcement learning policy.

\textbf{Supervised policy} is an end-to-end deep neural network that learns to map natural language user turns into system actions \cite{bingliu:2017}. 

\textbf{Reinforcement learning policy} is first initialized with the supervised policy model and further fine-tuned by interacting with an agenda-based user simulator (different from our simulators) \cite{bingliu:2017} to explore and generate more training data. Parameters of the neural network is updated using the REINFORCE algorithm \cite{williams92}. 
At each turn a small negative reward (such as -1) is given and at the end of each dialogue a large positive reward is given (such as 20) if the dialogue is successfully terminated.
As a result, RL policy tries to minimize the number of turns while achieving a high success rate.

These policies use natural language utterances as input and generate a sequence of system actions which are then converted into natural language using template-based generation. However, the user simulator only utilizes the dialogue acts and arguments of the system turns.
In our user simulator, we also leverage a template-based NLG to generate utterances as input to system policies where templates were sampled from a template set (about 10000) collected from crowdworkers.
Each policy is trained on a dataset different from ours and fixed.
To maintain a fair comparison, we first randomly generate 1000 user goals.
To evaluate policies and simulators, we generate a dialogue for all user goals by interacting each user simulator and policy pair.

\vspace*{-1ex}
\subsection{Dataset}
We use a task oriented dialogue corpus following \cite{shah18}.
At the beginning of each dialogue, a user goal is randomly generated and the dialogue is populated using a dialogue agent with finite state machine and an agenda based user simulator \cite{schatzmann07}.
Each dialogue also incorporates a user personality with different cooperativeness and randomness settings.
The training and testing datasets have 10,000 dialogues each.
Maximum number of dialogue turns is set to 20 and at each turn there are at most 3 sequences of dialogue actions with at most 5 slot-value pairs.
\vspace*{-1ex}
\subsection{Metrics}
We use three metrics to assess the successful interactions between user simulators and system policies: exact goal match, partial goal match, and dialogue length.

\begin{itemize}
	\itemsep 0ex
	\item A simulated dialogue has an \textit{exact goal match (EM)} score of 1 if the final slot-value pairs that system confirmed at the end of dialogue fully matches user's goal. 
	\item \textit{Partial goal match (PM)} score of a simulated dialogue is the number of correct slot-value pairs over the set of all possible slots in user goals.
	\item We also report numbers on \textit{dialogue length}, the average number of turns (both user and system) per dialogue, to assess the effect of different setups such as inducing noise.
\end{itemize}

Two response diversity metrics are also utilized to measure the diversity of user responses and the robustness of system policies: entropy and perplexity of the dialogue acts of the user responses per system response.

\subsection{Training Details}

We use randomly initialized word embeddings of size 150. We set state size of turn level and dialogue level RNN as 200. We train our models using Adam optimization method \cite{kingma14} with initial learning rate of 1e-3. We apply dropout with probability 0.5 and use mini-batches of size 32. We train our models for 10 epochs and choose the best model via validation.

\subsection{Task Completion and Dialogue Length Results}

In Table \ref{table:metrics}, we present our results related to task-completion and dialogue length metrics.
We used 10 different user simulation and system policy pairs.
In our earlier experiments, we also implemented a plain seq2seq user simulation model without explicit dialogue state tracking but we omitted from the paper due to having poor results.

\begin{table}
	
	\centering
	\scalebox{0.9}{
		\begin{tabular}{c|c|c|c|l}
			
			\cline{2-4}
			&\textbf{Exact} & \textbf{Partial} & \textbf{Dialogue } & \\
			&\textbf{Match (\%)} & \textbf{Match(\%)} & \textbf{Length} & \\ 
			\cline{1-4}
			
			\multicolumn{1}{ |c|  }{\multirow{2}{*}{HUS}} &  \cellcolor{lightgray} 75.67 &  \cellcolor{lightgray}94.3 &  \cellcolor{lightgray}12.03 & \textbf{SL} \\ \cline{2-4}
			\multicolumn{1}{ |c| }{} & 94.69 & 98.27 & 7.45 & \textbf{RL} \\  \cline{1-4}
			
			\multicolumn{1}{ |r|  }{\multirow{2}{*}{+ dialogue length}} &    \cellcolor{lightgray}86.1 & \cellcolor{lightgray}96.51 &  \cellcolor{lightgray}9.615 & \textbf{SL} \\ \cline{2-4}
			\multicolumn{1}{ |c| }{} & 94.33 & 98.2 & 7.076 & \textbf{RL} \\  \cline{1-4}
			
			\multicolumn{1}{ |c|  }{\multirow{2}{*}{VHUS}} &    \cellcolor{lightgray}82.52 &  \cellcolor{lightgray}95.69 &  \cellcolor{lightgray}11.8005 & \textbf{SL} \\ \cline{2-4}
			\multicolumn{1}{ |c| }{} & 95.53 & 98.43 & 7.803 & \textbf{RL} \\  \cline{1-4}
			
			\multicolumn{1}{ |c|  }{\multirow{2}{*}{HUSReg}} &    \cellcolor{lightgray}88.8 &  \cellcolor{lightgray}97.08 &  \cellcolor{lightgray}7.92 & \textbf{SL} \\ \cline{2-4}
			\multicolumn{1}{ |c| }{} & \textbf{96.19} & \textbf{98.56} & \textbf{6.878} & \textbf{RL} \\  \cline{1-4}
			
			\multicolumn{1}{ |c|  }{\multirow{2}{*}{VHUSReg}} &    \cellcolor{lightgray}91.90 &  \cellcolor{lightgray}97.67 &  \cellcolor{lightgray}8.0555 & \textbf{SL} \\ \cline{2-4}
			\multicolumn{1}{ |c| }{} & 95.98 & 98.52 & 6.905  & \textbf{RL} \\  \cline{1-4}
			
		\end{tabular}
	}
	\caption{Comparison of user simulation models and system policies using task completion and dialogue length metrics. Rows in gray show supervised learning based (SL) policy and rows in white show reinforcement learning based (RL) policy.
	}
	\label{table:metrics}
	\vspace*{-2ex}
\end{table}

\begin{table}
	\centering
	\scalebox{0.9}{
		\begin{tabular}{c|c|c|l}
			\cline{2-3}
			&\textbf{Entropy} & \textbf{Perplexity}& \\ \cline{1-3}
			\multicolumn{1}{ |c|  }{\multirow{2}{*}{HUS}} &   \cellcolor{lightgray}0.075 &  \cellcolor{lightgray}1.053 &  \textbf{SL} \\ \cline{2-3}
			\multicolumn{1}{ |c| }{} & 0.119	& 1.086 & \textbf{RL} \\  \cline{1-3}
			
			\multicolumn{1}{ |r|  }{\multirow{2}{*}{+ dialogue length}} &   \cellcolor{lightgray}0.091 &  \cellcolor{lightgray}1.065 & \textbf{SL} \\ \cline{2-3}
			\multicolumn{1}{ |c| }{} & 0.102 &	1.073 & \textbf{RL} \\  \cline{1-3}
			
			\multicolumn{1}{ |c|  }{\multirow{2}{*}{VHUS}} &  \cellcolor{lightgray}\textbf{0.284} & \cellcolor{lightgray} \textbf{1.218} & \textbf{SL} \\ \cline{2-3}
			\multicolumn{1}{ |c| }{} & 0.201 & 1.149  &  \textbf{RL} \\  \cline{1-3}
			
			\multicolumn{1}{ |c|  }{\multirow{2}{*}{HUSReg}} &   \cellcolor{lightgray}0.018 &  \cellcolor{lightgray}1.012  & \textbf{SL} \\ \cline{2-3}
			\multicolumn{1}{ |c| }{} & 0.0358 & 1.025 & \textbf{RL} \\  \cline{1-3}
			
			\multicolumn{1}{ |c|  }{\multirow{2}{*}{VHUSReg}} &  \cellcolor{lightgray} 0.211 & \cellcolor{lightgray}1.158 &  \textbf{SL} \\ \cline{2-3}
			\multicolumn{1}{ |c| }{} & 0.204 & 1.152  & \textbf{RL} \\  \cline{1-3}
		\end{tabular}
	}
	\caption{Comparison of user simulation models and system policies using response diversity metrics at the level of dialogue acts. 
	}
	\label{table:diversity}
	\vspace*{-3ex}
\end{table}

\subsubsection{Evaluating Dialogue Policies} When compared to SL policy, RL policy outperforms SL policy and generates remarkably higher task completion rates.
We observe that in the EM metric, the gap is larger than the PM metric.
One plausible explanation is that SL policy gets stuck in a local minima and can not recover some of the slot-value pairs. 
RL policy, on the other hand, is able to successfully produce the correct set of pairs.
RL policy also yields considerably fewer turns per dialogue across all user simulators.
When we decrease the complexity and effectiveness of user simulators (from bottom to top), the gap between RL and SL increases.
The performance of RL policy on EM and PM metrics is not affected by different user simulators; however SL policy shows a drop when a weaker user is present.
These observations indicate that RL is more robust to different types of users even when the responses are more stochastic (as in VHUS).
When we exclude dialogue length from HUS, SL performance drops due to its lack of robustness to a weaker user.
Note that these evaluation results obtained using fully automatic methods with a user simulator are in line with human evaluation results presented in~\cite{bingliu:2017}.

\subsubsection{Evaluating User Simulators}
We also compare user simulators and their components using the same set of policies and metrics.
When we incorporate dialogue length, HUS is able to untangle the effect of user types on dialogue length and other factors which lead to more successful and shorter conversations.
VHUS produces more diverse and more stochastic user responses which decreases the performance of SL policy and increases the average number of turns.
Although, RL policy can recover from these more unpredictable situations, the average length of dialogues increase; more than $0.7$ turns per dialogue.
Penalizing the user responses at each turn based on their divergence from the initial user goal generates more successful dialogues that are also shorter and more goal oriented.
One reason for the increase in EM and PM is that goal regularization tends to generate the shortest conversations possible and effectively alleviates the shortcomings of policies such as infinite loops.
When we augment HUSReg with the variational step, the performance of RL policy is not affected but the task completion rate of SL increases.
We also observe that average dialogue length does not increase which is attributed to the inverse effect of goal regularization to produce shorter conversations.

\begin{table}
	\centering
	\scalebox{0.9}{
		\begin{tabular}{|c|c|}
			\hline
			\textbf{Model} & \textbf{Average Score (Standard Deviation)}\\ \hline \hline
			Agenda-based & 4.56 (0.859)\\
			\hline
			HUS+dialogue length &   4.86 (0.545) \\ \hline
			VHUS &  4.88 (0.472) \\ \hline
			HUSReg &  4.88 (0.452) \\ \hline
			VHUSReg &  4.83 (0.594) \\ \hline
		\end{tabular}
	}
	\caption{Comparison of user simulator models using real user evaluation.}
	\label{table:user_eval}
	\vspace*{-3ex}
\end{table}

\subsection{Response Diversity Results}
Using the same set of simulated dialogues, in Table \ref{table:diversity}, we present our results related to response diversity metrics. Note that entropy and perplexity measures are computed at the level of dialogue acts to measure the ability of each approach in producing diverse responses, and the measures are expected to increase significantly when acts are converted to natural language utterances. 

\subsubsection{Evaluating Dialogue Policies}
RL policy produces slightly higher entropy and perplexity scores and generates richer dialogues over all user simulations except variational simulations.
We reason that inducing diversity into user responses is positively correlated with average dialogue length; hence RL policy will try to reduce very high uncertainty in user responses by balancing the number of turns and the total reward.
We also observe that the diversity of user responses drops when SL policy interacts with VHUSReg instead of VHUS while it is very similar in RL policy.
This similarity in RL policy performance can be associated to our reasoning above; RL policy tries to balance dialogue length and total reward which bounds the maximum diversity of user responses.

\subsubsection{Evaluating User Simulators}
VHUS and VHUSReg produces higher diversity scores over all metrics and system policies which confirms our hypothesis that introducing a variational step increases user response diversity.
The diversity scores on VHUSReg is smaller than VHUS in SL policy which is the result of the proposed goal regularization loss that controls the divergence of user responses from the initial goal.
HUSReg also generates dialogues that are more succinct with the smallest diversity scores and the smallest dialogue lengths compared to other user simulators.
When we incorporate dialogue length into HUS, we observe no change in diversity scores while improving task completion and dialogue length performances.

\subsection{Evaluating with Real Users}
We presented 100 dialogue subset of each dataset to crowdworkers for human evaluation.
All turns of these dialogues were transformed into natural language by template based generation and  annotated by 3 crowdworkers in a scale from 1 to 5 in terms of the clarity and appropriateness of that turn in the context of the conversation. 
Table \ref{table:user_eval} shows the turn-based average scores.
We observe that all user simulation models generate highly successful turns.
We also compare our user simulation models to a hand-crafted agenda-based user simulation model \cite{schatzmann07}.
Our models have higher average user scores with less standard deviation compared to agenda-based user simulation.
One particular explanation for this difference is that agenda-based simulation can generate out of context turns deviating from the dialogue history.

\section{Conclusions}
We studied the problem of modeling users in task oriented dialogues and proposed several end-to-end model architectures that are able to successfully interact with different system policies.
Via systematic interaction between our user simulators and state-of-the-art system policies, we provide new insights into evaluating task-oriented dialogues.
We show that RL policies are more focused towards understanding hidden intents of users and more robust to variations in user responses.
By incorporating a variational step into HUS, we are able to introduce meaningful diversity into our models.
We introduced a new goal regularization approach that effectively reduces average number of turns while increasing success rate.
Finally, the proposed hybrid model strikes a good balance between generating diverse responses and successful user turns.

\bibliographystyle{IEEEbib}
\bibliography{slt2018}

\end{document}